# Constructing Reference Sets from Unstructured, Ungrammatical Text

**Matthew Michelson**                                    MMICHELSON@FETCH.COM
*Fetch Technologies*
*841 Apollo St, Ste 400*
*El Segundo, CA 90245 USA*
**Craig A. Knoblock**                                    KNOBLOCK@ISI.EDU
*University of Southern California*
*Information Sciences Institute*
*4676 Admiralty Way*
*Marina del Rey, CA 90292 USA*

## Abstract

Vast amounts of text on the Web are unstructured and ungrammatical, such as classified ads, auction listings, forum postings, etc. We call such text "posts." Despite their inconsistent structure and lack of grammar, posts are full of useful information. This paper presents work on semi-automatically building tables of relational information, called "reference sets," by analyzing such posts directly. Reference sets can be applied to a number of tasks such as ontology maintenance and information extraction. Our reference-set construction method starts with just a small amount of background knowledge, and constructs tuples representing the entities in the posts to form a reference set. We also describe an extension to this approach for the special case where even this small amount of background knowledge is impossible to discover and use. To evaluate the utility of the machine-constructed reference sets, we compare them to manually constructed reference sets in the context of reference-set-based information extraction. Our results show the reference sets constructed by our method outperform manually constructed reference sets. We also compare the reference-set-based extraction approach using the machine-constructed reference set to supervised extraction approaches using generic features. These results demonstrate that using machine-constructed reference sets outperforms the supervised methods, even though the supervised methods require training data.

## 1. Introduction

There are vast amounts of unstructured, ungrammatical data on the Web. Sources of such data include classified ads, auction listings, and bulletin board/forum postings. We call this unstructured, ungrammatical data "posts." Figure 1 shows example posts, in this case a set of classified ads for cars from the Craigslist site. We consider posts to be unstructured because one cannot assume that the ordering of the terms will be consistent from post to post. For instance, in Figure 1 we cannot assume that the car make (e.g., Audi) will precede the car model (e.g., A4). Further, we consider posts ungrammatical because they would not yield a useful grammatical parse (it would yield all nouns almost exclusively).

Although posts are unstructured and ungrammatical, they are full of useful information. The posts of Figure 1 contain information about cars such as the different varieties of cars for sale, their price, etc. Therefore, analyzing posts as a corpus of information can yield a number of insights.

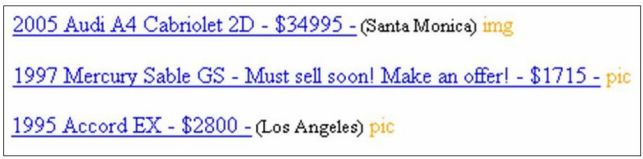

Figure 1: Example posts for cars

In this paper we focus on analyzing posts to build "reference sets," which are relational tables of entities and their attributes. For instance, a reference set about cars would include attributes such as a car make (e.g., Mercury), a car model (Sable), and a car trim (GS). Reference sets can be used as background information for a number of tasks, and in particular, there has been recent interest on using reference sets as background knowledge for information extraction (Michelson & Knoblock, 2007, 2008; Mansuri & Sarawagi, 2006; Cohen & Sarawagi, 2004). For completeness, we describe previous work on reference-set-based information extraction in more detail in Section 2.

Our goal is to exploit the dense information content of posts to construct reference sets with as little human intervention as possible. Posts are numerous, freely available, and generally have a large number of useful attributes packed into short strings. Further, posts sometimes cover more obscure entities than one might find in general Web text (another possible corpus for mining a reference set). For instance, there are many categories of items on auction sites that are specific and particular. Such items may not be mentioned frequently in Web text outside of the posts. Therefore, posts are an attractive data source for building reference sets.

We contrast our approach to manually constructing reference sets, which poses a number of challenges. Most importantly, a user must discover the correct source(s) from which to build the reference set. While reference sets can be built by extracting information from semi-structured websites using "wrappers" (Muslea, Minton, & Knoblock, 2001; Crescenzi, Mecca, & Merialdo, 2001), the challenge is not in parsing the data, but rather in choosing the correct sources. There are often cases where it is difficult to find sources that enumerate the reference set of interest, and therefore building the reference set itself becomes a challenging problem of finding multiple sources to use, aggregating their records, and taking care of duplicates. For instance, as we demonstrate in our results, it is difficult to find single sources that enumerates very specific attributes such as the model numbers of laptops. There is a Dell Inspiron 1720, an Inspiron E1405, and a 2650, just to name a few. This happens across IBM laptops, Dell laptops, HP laptops, etc. To create a reference set that encompasses all of the laptop model numbers would require finding and scraping a multitude of websites. Yet, this is a useful attribute for extraction since the difference between an Inspiron 1720 and E1405 is important to the end user. However, posts are generally an aggregation of the items people are interested in, so by using posts for the reference set one can discover a more exhaustive list.

Further, the data for reference sets is constantly changing. For example, again considering the laptop domain, new laptop model numbers are released every few months as new hardware improvements are made. Therefore, even if a comprehensive laptop reference set is built from multiple sources, it will become stale after just a few months, and the reference set will need to be updated each time any of the sources is updated with new laptops. This creates an additional challenge for creating these reference sets from wrapped sources because the sources must be constantly monitored for changes to assure that the reference set reflects those changes.

Lastly, as we stated above, much of this work is motivated by using reference sets as background knowledge to aid information extraction (Section 2). However, even if a user can extract a reference set from Web sources using wrappers, there is no guarantee that the reference set is appropriate given the data for extraction. Concretely, assume the goal is to extract car information from Craigslist classifieds such as those in Figure 1, and assume a user built a reference set about American cars using Web sources. This reference set is not useful if the posts are for cars from outside the United States. We call this mismatch between the reference set and the posts a "coverage" problem because the entities in the posts are not covered by the records in the reference set.

Such coverage mismatches also happen when the data is constantly changing, such as the laptops described above. In this case, if websites for new laptops are used to create the reference set, there might be a mismatch for the used laptops that are generally sold in classified ads (in fact, we demonstrate just this effect in our results). Therefore, there are a number of advantages to building reference sets from posts instead of Web sources, although of course these could be complementary processes.

This paper elaborates on our previous work on building reference sets from posts (Michelson & Knoblock, 2009). First, we describe our previous method in more detail. Second, this previous method exploits a small amount of background knowledge to build the reference set, and so in this paper we also significantly extend the algorithm to handle the special case when no background knowledge about the source is known. We demonstrate that for even a totally unknown source we can still construct a reference set that is useful for extraction.

The rest of the paper is as follows. Section 2 describes in more detail the process of using reference sets for information extraction. Section 3 then describes our method for constructing reference sets from the posts themselves. Section 4 presents our experiments and analyzes the results. Section 5 discusses the related work, and Section 6 presents our conclusions and future work.

## 2. Previous Work using Reference Sets for Information Extraction

We motivate our work on constructing reference sets by putting it in the context of discovering reference sets for use in information extraction. In particular, we focus on the task of extracting information from posts using reference sets, as they have been shown previously to aid this task quite a bit (Michelson & Knoblock, 2005, 2007). Information extraction is the task of parsing the desired attributes from some text (posts in this case). Using the example posts of Figure 1, after extraction, we would like the posts of Figure 1 to be separated into a make attribute, a model attribute, and a trim attribute. Specifically, the third post could be annotated with extractions such as MAKE={Honda} (which is implied!), MODEL={Accord}, and TRIM = {EX}. Once done, this would allow for structured queries and integration over the posts, using the extracted attributes instead of the full text of the post.

Given that posts do not conform to English grammar, Natural Language Processing approaches to extraction would not work on posts (Ciravegna, 2001). Further, since the posts are unstructured, wrapper methods would not work either (Muslea et al., 2001; Crescenzi et al., 2001). Instead, recent work showed that extraction from posts benefits from exploiting reference sets (Michelson & Knoblock, 2005, 2007).

Previous work on exploiting reference sets for extraction uses a two step process, shown in Figure 2. In the first step, the algorithm discovers the best matches between each post and the tuples in the reference set. Then, during the second step, the algorithm performs information extraction by comparing the tokens in each post to the attributes of the matches from the reference set. For example, the token "Accord" would be extracted as a model since it would match the model attribute of the matching tuple from the reference set. The key idea is that without labeled data, the system can perform information extraction by relying solely on the information in the reference set, rather than some grammatical or structural patterns. Recent work shows how to perform reference-set-based extraction automatically (Michelson & Knoblock, 2007) or using techniques from machine learning (Michelson & Knoblock, 2008).

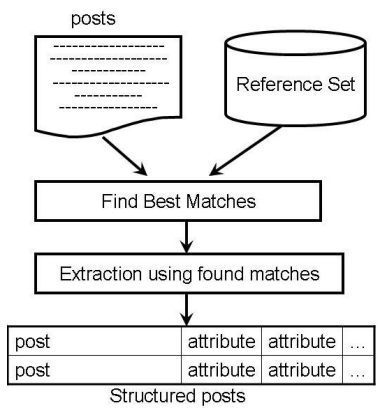

Figure 2: Reference-set based extraction

This previous work on reference-set-based extraction assumes the user has a manually constructed reference set to supply to the algorithm. This is a strong assumption and has many potential difficulties as we stated above. Therefore, this helps motivate the approach we take in this paper, building reference sets from posts, with little (or no) human intervention. The method in this paper does not rely on any labeled data, and may only require a small amount of background knowledge to produce clean and useful reference sets. These machine-constructed reference sets can then be applied by plugging them into a reference-set-based extraction framework for automatic extraction from posts.

Within the context of information extraction, we emphasize that there are two main benefits for using the posts themselves to construct the reference set. First, the challenge of discovering sources from which to manually build the reference set is overcome since the posts themselves can be used. Second, since the reference set is constructed from the posts, it will have overlapping coverage by definition. As we show in our experiments, our approach is especially useful for covering those attributes that are particularly difficult to cover with manual reference sets, such as laptop model numbers, which constantly change and are hard to enumerate.

We note that although we motivate this work on building reference-sets in the context of information extraction, these machine constructed reference sets are useful for many other tasks that require background knowledge. We can use them as structured entities to fill in taxonomies, help construct ontologies for the Semantic Web, or simply see what tuples of a reference set might exist in the posts, which can help in topic classification or query formulation. We emphasize that the information extraction task as described here is one application of these reference sets, and it is a useful proxy for comparing the utility of various reference sets (as we do in our experiments). However, the work described in this paper focuses on the specific task of reference set construction, rather than extraction which is described elsewhere.

## 3. Reference Set Construction

The intuition for constructing reference sets from posts is that reference set tuples often form hierarchy-like structures, which we call "entity trees." Consider the simple reference set of three cars in Figure 3. Each tuple in the reference set of cars has two attributes, a make and a model. The model is a more specific version of the make (i.e., an Accord is a Honda), and so the Accord attribute value is a child node to the Honda value. The same can be said of the Civic value, and so we can generate an entity tree rooted on the value Honda, with two children: Civic and Accord. The same argument applies to turn the Ford tuple into its own entity tree. However, the entity trees rooted on Honda and Ford are disjoint, since they do not share any ancestors. So a reference set is really a forest of entity trees. So, once an algorithm constructs the set of entity trees, it can then traverse them, outputting a tuple for each path from root to leaf, and the union of all of these tuples creates a reference set. Therefore, to construct a reference set from posts, the goal is really to build the forest of entity trees from the posts.

Our general approach to building reference sets from posts decomposes into two high-level steps, shown in Figure 4. The first step breaks the posts into bigrams. As an example, the post, "Honda civic is cool" yields the bigrams: {Honda, civic}, {civic, is}, etc.[1] Creating bigrams is a pre-processing step, since the second step of the algorithm takes the full set of bigrams as input and generates the reference set.

The second step of our algorithm has three sub-components. First, the algorithm generates an initial set of entity trees based upon the bigrams (shown as step 2.1). Next, it iteratively adds new attributes to the entity trees that are "general token" attributes, which we define below (step 2.2). Finally, the algorithm traverses each entity tree from root to leaf, outputting each path to generate the reference set (step 2.3). We discuss our methods for creating entity trees and discovering general tokens in detail in the following subsections.

---

1. Note that our algorithm only considers ordered bigrams, rather than all combination of token pairs. This is done for efficiency, since we measured an average post length of 8.6 tokens across thousands of posts, which would generate more than 40,320 possible token pairs to check, per post, if all pairs are considered.

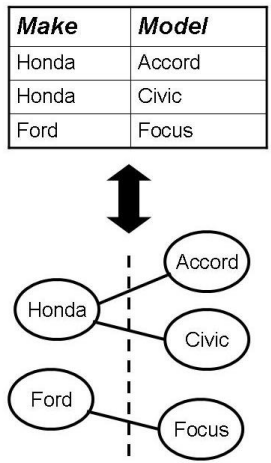

Figure 3: A reference set and its entity trees

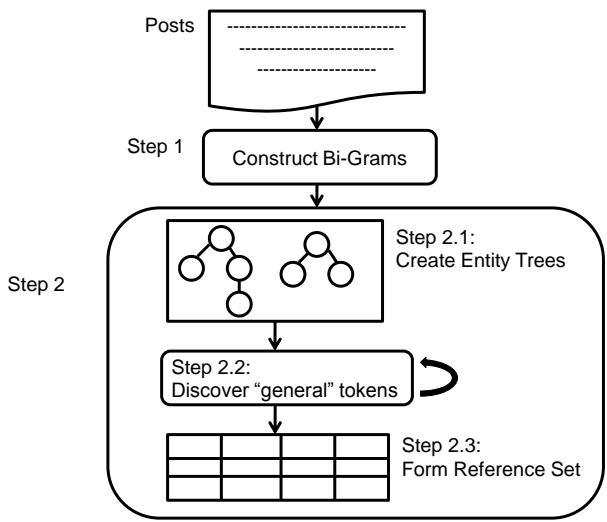

Figure 4: Constructing reference sets from posts

### 3.1 Creating Entity Trees

As stated above, the first step in our approach converts the set of input posts into a set of bigrams. To build entity trees from the bigrams we use a modified version of the Sanderson and Croft (1999) heuristic for finding token subsumptions, with the notion that if token $x$ subsumes $y$, then $y$ is a child node under $x$ in the entity tree that contains $x$. We define the rule for subsumption given terms $x$ and $y$ as:[2]

$$x \text{ SUBSUMES } y \rightarrow P(x|y) \geq 0.75 \text{ AND } P(y|x) \leq P(x|y)$$

As an example, consider the four posts shown in Table 1. If we consider the token pair "Honda" with "Civic" we see that the conditional probability of Honda given Civic ($P(\text{Honda}|\text{Civic})$) is 1.0, while $P(\text{Civic}|\text{Honda})$ is 0.5 (since Honda occurs in four posts while Honda occurs with Civic in two). Therefore, the subsumption rule fires, and Civic becomes a child node to Honda. Similarly, Accord becomes a child of Honda, resulting in the entity tree of Figure 3.

Table 1: Four example posts

| Honda civic is cool |
|---|
| Look at cheap Honda civic |
| Honda accord rules |
| A Honda accord 4 u! |

Since we only consider the ordered bigrams, building entity trees based on the Sanderson and Croft heuristic requires the assumption that the order in the entity tree is reflected in the posts. That is, users tend to order the attributes from more general (e.g., car makes) to more specific (e.g., car models). Also, this ordering needs to hold in at least enough of the posts to make the subsumption rule fire.

Yet, once our approach builds the entity trees, it constructs useful reference sets which can be used effectively for extraction, as shown in our experiments. Therefore, our approach leverages the little ordering it does find in the bigrams to build the reference set, which we can then use to extract values from posts where the ordering assumption does not hold. Further, given that our approach finds effective entity trees at all reflects the notion that users do tend to order attributes from general to specific. If this were not the case, the discovered entity trees would have little utility for extraction later.

The Sanderson and Croft heuristic above is defined for single tokens $x$ and $y$, yet not all attribute values are unigrams. Therefore, to handle bigrams, we add the constraint that if $x$ subsumes $y$ and $y$ subsumes $x$, we merge $x$ and $y$ into a single node (attribute value). For instance, given attribute values "Crown" and "Victoria" if "Crown" SUBSUMES "Victoria" and "Victoria" SUBSUMES "Crown" we merge them into a single value "Crown Victoria" (which is subsumed by the common parent "Ford"). We note this bigram was actually merged using our approach. To extend this to n-grams, one simply checks all token pairs against each other for subsumption.

---

2. Note, we require terms $x$ and $y$ to co-occur at least once for this rule to be considered.

In the reference sets constructed by our algorithm for our experiments, 5.49% of the attribute values are n-grams containing more than a single token, which is a large enough percentage to validate including the merge heuristic in our approach. Therefore, our technique is not solely applicable to the case where reference set values are single tokens. However, there are cases where our merging heuristic does not perform well, which we discuss in more detail in our discussion in Section 4.3.

The directionality imposed by the Sanderson and Croft heuristic is important for our approach. Specifically, the Sanderson and Croft heuristic uses conditional probabilities which imposes a directionality on the relationship between the tokens and allows them to form subsumption relationships. That is, if $x$ subsumes $y$, then $x$ is a parent of $y$ based on the directionality imposed by the conditional probability. We need this directionality because the entity trees must have a relative subsumption ordering to align the trees into a reference set. More specifically, using the example from Figure 3, when building the reference set from the entity trees, we know that the columns of the reference set will be aligned based on their positions in the entity tree. So, the roots of entity tree form the leftmost column of the reference set table (Honda, Ford), their children (Honda, Accord, Focus) form the next right column, etc., down to the leaves of the entity trees. So, when tracing the paths from root to leaf, we can be sure that the tuples outputted from disjoint trees will still align correctly in the columns of the reference set because of the ordering imposed by the directionality.

This is in contrast to other probabilistic measures of term co-occurrences such as Pointwise Mutual Information (PMI).[3] Specifically, PMI is symmetric and therefore while it provides a strong measure of term relationships, it is unclear how to order terms based on this measure, and therefore, even if one were to use such a symmetric relationship to find reference set tuples, it is unclear how to align disjoint tuples (e.g., Honda and Ford tuples) based on this metric since it does not impose a relative ordering of the attributes. For this reason, we require an asymmetric measure.

## 3.2 Discovering General Tokens

Once we have constructed the initial entity trees, we then iterate over the posts to discover possible terms that occur *across* trees. Specifically, subsumption is determined by the conditional probabilities of the tokens, but when the second token is much more frequent than the first token, the conditional probability will be low and not yield a subsumption. This occurs when the attribute value appears across entity trees (reference set tuples). Since the second term occurs more frequently than the first across tuples, we call this the "general" token effect.

An example of this general token effect can be seen with the trim attribute of cars. For instance, consider the posts in Table 2 which show the general token effect for the trim value of "LE." These posts show the "LE" trim occurring with a Corolla model, a Camry model, a Grand AM model, and a Pathfinder. Since LE happens across many different posts in many varying bigrams, we call it a "general" token, and its conditional probability will never be high enough for subsumption. Thus it is never inserted into an entity tree.

---

3. For terms $x$ and $y$, PMI is defined as $PMI(x, y) = \log \frac{p(x,y)}{p(x)p(y)}$

Table 2: Posts with a general trim token: LE

| |
|---|
| 2001 Nissan Pathfinder LE - $15000 |
| Toyota Camry LE 2003 — 20000 $15250 |
| 98 Corolla LE 145K, Remote entry w/ alarm, $4600 |
| 1995 Pontiac Grand AM LE (Northern NJ) $700 |

To compensate for this "general token" peculiarity, we iteratively run our subsumption process, where for each iteration after the first, we consider the conditional probability using a set of the first tokens from bigrams that all share the common second token in the bigram. Note, however, this set only contains bigrams whose first token is already a node in an entity tree, otherwise the algorithm may be counting noise in the conditional probability. This is the reason we can only run this after the first iteration. The algorithm iterates until it no longer generates new attribute values.

To make this clear, consider again the 'LE' trim. This is a possible cross-tree attribute because it would occur in the disjoint subtrees rooted on CAMRY, COROLLA, and PATHFINDER. By iterating, the algorithm considers the following conditional probability for subsumption, assuming the models of Camry, Corolla and Pathfinder have already been discovered:

$$P(\{CAMRY \cup COROLLA \cup PATHFINDER\}|LE)$$

Now, if this conditional probability fits the heuristic for subsumption, then LE is added as a child to the nodes CAMRY, COROLLA and PATHFINDER in their own respective entity trees. Iterating is important for this step since our method only tests a general token for subsumption if the terms it occurs with as a bigram are already in some entity trees. So, if LE occurred with other tokens, but none of them were in an entity tree already, our method ignores them as noise. Our approach must iterate since new general tokens are only considered if they occur with an attribute in an entity tree, so our approach may discover a new general token and add it to an entity tree, which in turn allows the approach to discover another new general token. In our experiments we describe the effects of iterating versus not. The final step of our approach from Figure 4 then turns the entity trees into a reference set by tracing down the paths of the trees, outputting the nodes as columns in the reference set tuples.

However, blindly applying the above subsumption method can lead to noisy entity trees. A common occurrence in auction listings, for instance, is to include the term "Free Shipping" or "Free Handling." If such phrases occur frequently enough in the posts, the subsumption rule will fire, creating an entity tree rooted on Free with the children Shipping and Handling. Clearly this is a noisy tree and when used it would introduce noisy extractions. Therefore, the following two subsections describe different approaches to handling noise in the process of constructing entity trees.

### 3.3 Seed-Based Reference Set Construction

Our first approach to dealing with noise exploits a small amount of background knowledge, called "seeds," to focus the construction of the entity trees. Specifically, we use the method of Figure 4 to build entity trees, with the added constraint that each entity tree must be rooted on a given seed value. If we only gave "Honda" as a seed, then only one entity tree rooted on Honda would be constructed. Even if other entity trees are discovered, they are discarded. It is easy to discover an exhaustive list of seeds on the Web and including too many seeds is not a issue as our algorithm simply removes any entity tree that consists solely of a root from the constructed reference set (i.e., a singleton set).

One of the key intuitions behind our approach is that the set of root nodes for entity trees are generally much easier to discover online than nodes farther down the trees. For instance, if we consider laptops, the manufacturers of laptops (the roots) are fairly easy to discover and enumerate. However, as one traverses farther down the entity trees, say to the model numbers, it becomes hard to find this information. Yet, just this small set of seeds is enough to improve the process of reference set construction substantially, as we show in the results where we compare against reference sets constructed without the seed-based constraint. Also, importantly, the attributes farther down the tree change more over time (new model numbers are released often), while the seeds infrequently change (there are few new computer manufacturers). So, when using the reference set for information extraction, coverage becomes less of an issue when only considering the roots versus all of the attributes in a reference set tuple.

In this manner we construct a reference set directly from the posts, using the seeds to constrain the noise that is generated in the final reference set. Table 3 describes our full algorithm for constructing entity trees using seeds, which are then turned into a reference set by outputting a tuple for each path from root to leaf in each tree.

## 3.4 Locking-Based Reference Set Construction

The seed-based method handles noise by exploiting a small amount of background knowledge. Here we describe a second approach to dealing with the noise for the case where the seeds are impossible to discover or too costly to find and use.

This approach revolves around a "locking" mechanism. The seeds in the seed-based method constrain the possible entity trees by limiting the attributes that can become roots of the entity trees. This, in turn, leads to cleaner and more useful reference sets. Therefore, when lacking seeds, the goal should be to introduce a constraining mechanism that prevents noise from being introduced into the entity trees. Our approach is to lock the levels of the entity trees at certain points, such that after locking no new attribute values can be introduced at that level for any of the entity trees.

Since many of the attributes we discover are specifications of more general attributes (such as car models specify makes), there is a point at which although we may be discovering new values for the more specific attributes (car models), we have saturated what we can discover for the parent attribute (car makes). Further, once we saturate the parent attributes, if the algorithm does introduce new values representing new parents, they are likely noise. So, the algorithm should "lock" the parent attribute at the saturation point, and only allow the discovery new attributes that are below the level of locked attribute in the hierarchies.

Consider the example shown in Figure 5. At iteration $t$ the algorithm has discovered two entity trees, one rooted with the car make Ford and one rooted with the car make Honda. Each of these makes also has a few models associated with them (their children).

Table 3: Entity tree construction using seeds

---

MineReferenceSet(Posts $P$, Seeds $S$)

---

```
/* First, break posts into bigrams */
Bigrams B ← MakeBigrams(P)

/* Build the entity trees rooted on the seeds */
AddedNodes N ← {}
For each {x, y} ∈ B            /* {x, y} is a bigram */
    If x ∈ S                   /* check x is a seed */
        If x SUBSUMES y
            y is child of x in entity tree
            N ← N ∪ y

/*Find all children's children, and their children, etc.*/
While N is not null
    For each {s, t} ∈ B
    where s ∈ N
        N ← N - s
        If s SUBSUMES t
            t is child of s in tree
            N ← N ∪ t

/* Iterate to discover ``general'' token nodes */
/* Start with unique nodes already in the entity trees */
AllEntityNodes ← UniqueList(All Entity Trees)
/* Keep iterating while find new terms */
FoundNewTerms ← true
While FoundNewTerms is true
    FoundNewTerms ← false
    /* Consider terms {p₀, …, pₙ} in entity the trees
       that all form bigrams with non-entity tree term q */
    For each ( {p₀, …, pₙ}, q) s.t. {p₀, …, pₙ} ⊂ AllEntityNodes
        If {p₀, …, pₙ} SUBSUMES q        /* consider P(⋃pᵢ|q) */
            q becomes child of each pᵢ in trees
            AllEntityNodes ← AllEntityNodes ∪ q
            FoundNewTerms ← true
```

---

The bottom of the figure shows some future time (iteration $t+y$), at which point the system decides to lock the make attribute, but not the model and trim. Therefore, the algorithm can still add car models (as it does with the Taurus model to the Ford make) and car trims (as with the LX trim for the Civic model). However, since the make attribute is locked, no more makes attributes are allowed, and therefore the hierarchy that would have been rooted on the make "Brand" with model "New" (which is noise) is not allowed. This is why it is shown on the right as being crossed out.

In this manner, the locking acts like a pre-pruner of attribute values in a similar spirit to the seeds. The intent is that the algorithm will lock the attributes at the right time so as to minimize the number of noisy attributes that may be introduced at later iterations. This works because noise is often introduced as the algorithm iterates, but not in the beginning. In our example, instead of post-pruning away the hierarchy rooted on "Brand," the algorithm instead prevented it from being added by locking the make attribute.

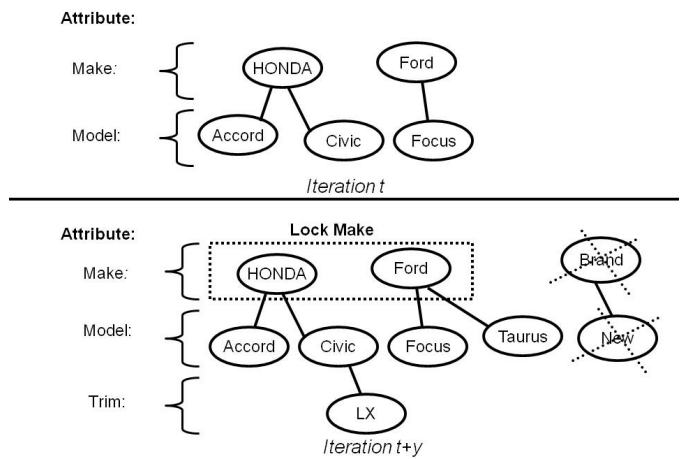

Figure 5: Locking car attributes

The assumption is that for deeper levels of the tree, more posts will need to be examined because the attributes represented in these levels of the trees are rarer. However, the higher the level of the tree, the more common the attribute value and therefore the fewer posts will need to be examined. This is an assumption that we will exploit to justify locking in a top down manner. That is, we can lock the roots of the entity trees before their children because we assume that we will need to see fewer posts to discover the roots than the children. This assumption carries down the tree, such that we lock the leaves last because we need to see the most posts to generate the leaves. So, the locking approach terminates if all attributes (i.e., all levels of the entity trees) become locked, since then there is no point in processing any more posts after that.

Based on our assumption, we model the locking mechanism as a requesting service. Our assumption is that we need to look at fewer posts for higher up in the trees, and that by doing so we help eliminate noise (justified in our experiments by comparing locking to not locking). So rather than process all of the posts at once, we instead have the algorithm request batches of posts and process them, building the entity trees using the approach of Figure 4. After receiving the next batch of posts, the algorithm builds up the new set of entity trees, compares them to the previously discovered ones, and the algorithm decides whether to lock a level. If it does not lock, it requests more posts to examine. It keeps requesting posts to examine until all levels of the entity trees are locked. We note, however, that as each batch comes in for processing, it is combined with all of the previously seen posts. This way, there is a gradual build up of the number of posts the system examines for a given level, and so in this sense the algorithm iterates because it examines the previous posts each time it receives a new batch.

Therefore, we can leverage the notion that the locking process requires continuous requests for more posts from the user until it locks all attributes. Essentially, at each request, the machine compares what it can discover using the given posts to what it can discover using the previous set of given posts (note that the newly requested set supersedes the previous set). For example, the algorithm may start by examining 100 posts to build a reference set. Then, if not all levels in the entity trees are locked, the algorithm requests 100 more posts and now analyzes these 200 posts to build the reference set. The algorithm compares what it can discover by using the first 100 posts to what it can discover using the combined 200 posts. If the algorithm thinks it cannot discover more useful attributes from the 200 posts than the 100 posts, then it locks certain attributes (Again, note that the 100 posts are a subset of the 200 posts).

To do this comparison for locking, the algorithm compares the entropies for generating a given attribute (e.g., member of a given level in an entity tree), based upon the likelihood of a token being labeled as the given attribute. So, in the cars domain we would calculate the entropy of a token being labeled a make, model, or trim (note that these label names are given post-hoc, and the machine simply regards them as attribute$_1$, attribute$_2$, etc.). For clarity, if we define $p_{make}(x)$ as the probability of arbitrary token $x$ being labeled as a *make* attribute (e.g., falling into the level of an entity tree associated with make), then we define the entropy, $H(make)$ as:[4]

$$H(make) = - \sum_{x \in tokens} p_{make}(x) * \log(p_{make}(x))$$

So, for any given attribute $A$, we use $p_A(x)$ and define H(A) as:

$$H(A) = - \sum_{x \in tokens} p_A(x) * \log(p_A(x))$$

The entropy of labeling a particular attribute can be interpreted as the uncertainty of labeling tokens with the given attribute. So, as we see more and more posts, if the entropy does not change from seeing 100 posts to seeing 200 posts, then the uncertainty in labeling that attribute is steady so there is no need to keep mining for that attribute in more posts. However, we cannot directly compare the entropies across runs, since the underlying amounts of data are different. So, we use normalized entropy instead. That is, for attribute $A$ (e.g., *make*), given $N$ posts, we define the normalized entropy H(A)$_N$:

$$H(A)_N = \frac{H(A)}{\log N}$$

Although the entropy provides a measure of uncertainly for token labels, it does not provide a sufficient comparison between runs over varying numbers of posts. To provide an explicit comparison, we analyze the percent difference between the normalized entropies across runs. For instance, using attribute $A$, we would compare the entropies from runs across 100 posts and 200 posts (defined as H(A)$_{100}$ and H(A)$_{200}$ respectively), by computing the percent difference between them:

$$PD(H(A)_{100}, H(A)_{200}) = \frac{|H(A)_{100} - H(A)_{200}|}{\frac{1}{2}(H(A)_{100} + H(A)_{200})}$$

If this value is a minimum (ideally 0), we know we can lock that attribute at 100 posts, since using the 200 posts did not yield more information as the entropies are essentially the same (i.e., the uncertainty is steady). So, the algorithm locks an attribute when it finds the minimum percent difference between entropies across runs for that attribute. This minimum is found using a greedy search over the previously calculated PD values. Table 4 summarizes the above technique for locking the attributes when mining a reference set.

---

4. This assumes we have built a reference set from which to compute the probabilities $p_{make}(x)$.

Table 4: Locking attributes

LockingAttributes(Attributes A, Posts $P_i$, Posts $P_j$,
ReferenceSet $RS_i$, ReferenceSet $RS_j$, LockedAttributes L)
/* $P_i$ is the first $i$ set of posts */
/* $P_j$ is the next set of $j$ posts, such that $P_i \subset P_j$ */
/* $RS_i$ and $RS_j$ are the reference sets for $P_i$ and $P_j$
 respectively and are used with the posts to compute likelihoods $p_A(x)$ */
/* L is the set of previously locked attributes */

for each attribute $a \in A$
if $a$ is not locked
    /* Compute entropies $H(a)_i$ and $H(a)_j$ as defined above */
    $H(a)_i \leftarrow$ Compute entropy for $a$ given $P_i$ and $RS_i$
    $H(a)_j \leftarrow$ Compute entropy for $a$ given $P_j$ and $RS_j$
    If $PD(H(a)_i, H(a)_j)$ is a minimum over previous values AND
    Parent($a$) $\in$ L /* Parent of $a$ is locked */
        Lock($a$)
        L $\leftarrow$ L $\cup$ $a$
return L

There are two small items to note. First, we add a heuristic that an attribute may only lock if its parent attribute is already locked, to prevent children from locking before parents. Second, although the above discussion repeatedly mentions the machine requesting more posts from a user, note that the algorithm can automatically request its own posts from the sources using technology such as RSS feeds, alleviating any human involvement beyond supplying the URL to the source.

Therefore, by using the above technique to lock the attributes as we go, even if we are generating new children attribute values, the parents will be locked, which acts as an automatic pruning method. Further, we now know when to stop asking for posts, so we have a stopping criteria for the number of posts that we need to run the mining algorithm. This stopping criteria is important since the assumption is that the algorithm has exhausted possible reference set membership only when it locks all levels. Note, this assumption could be violated if the algorithm is not supplied with enough posts (since it would never lock all levels).

Table 5 ties all of the aspects together describing our "Iterative Locking Algorithm" (ILA) for mining reference set tuples from posts without seeds. Essentially, the approach is the same as the seed-based method: for each batch of posts, the algorithm constructs the entity trees by scanning the data and iterating for the "general" tokens. The key difference is that ILA does not constrain the roots of the entity trees to be seeds, but instead uses locking to block attributes from being added to the trees. Further, this locking approach processes batches of posts at a time, so the flow is different in that it requests a batch of posts, builds/refines the entity trees, tests the trees for the locking conditions, and then proceeds with the next batch of posts.

Table 5: ILA method for mining reference sets

| |
|---|
| MineReferenceSet(Posts, $x$, $y$) |
| # $x$ is the number of posts to start with |
| # $y$ is the number of posts to add each iteration |

Posts $P_x \leftarrow$ GetPosts($x$)
ReferenceSet RS$_x \leftarrow$ BuildReferenceSet(P$_x$)
`/* Algorithm from Table 3, except`
`ignore constraint that tree roots are seeds */`
finished $\leftarrow$ false
LockedAttributes L $\leftarrow \{\}$
while(finished is false)
    $x \leftarrow x + y$
    Posts $P_y \leftarrow$ GetPosts($x$)
    ReferenceSet RS$_y \leftarrow$ BuildReferenceSet(P$_y$, L)
    `/* Same as in Table 3, except`
    `only add nodes to unlocked levels of the trees */`
    Attributes A $\leftarrow$ GetAttributes(RS$_y$)
    `/* GetAttributes returns columns of found reference set */`
    L $\leftarrow$ L $\cup$ LockingAttributes(A, P$_x$, P$_y$, RS$_x$, RS$_y$, L)
    If $|$L$|$ equals $|$A$|$ `/* Same size, so all attributes are locked */`
        finished $\leftarrow$ true
    P$_x \leftarrow$ P$_y$
    RS$_x \leftarrow$ RS$_y$

## 4. Experiments

The goal of this research is to to construct reference sets for tasks such as information extraction from posts. However, directly comparing the reference sets constructed in various ways has a number of problems. First, it is unclear how to directly compare reference sets quantitatively. This problem has been noted elsewhere as well in the context of comparing hierarchies (e.g., Bast, Dupret, Majumdar, & Piwowarski, 2006). For instance, there are not clear measures on how to define similarities when comparing hierarchies. Second, without context, it is hard to judge the reference sets. That is, one reference set may be huge and comprehensive, which is high utility for a task such as ontology construction, but might be of little use for extraction due to coverage mismatches. Another reference set may be quite noisy (therefore bad for ontology construction), but actually be preferred for extraction since its coverage is better.

Therefore, instead of measuring the "goodness" of reference sets directly, we instead put them in the context of an information extraction task by using them for reference-set-based extraction from posts. Then we can compare the extraction results and use these results as a proxy for the utility of the reference set. The assumption is that the extraction results reflect the reference set's utility since a very noisy reference set will lead to poor extraction, as would a reference set with poor coverage to the posts.

In the following subsections we test our two different approaches to constructing reference sets: our seed-based approach and our locking based approach (ILA).

### 4.1 Experiments: Seed-based Approach

For our first experiment, we test the effectiveness of using the seed-based approach for building reference sets. For this experiment, we compare our seed-based approach to both full, manually constructed reference sets extracted from online sources (which we call the "manual" approach) and to a version of the seed-based approach that does not constrain the entity trees to be rooted on seed values (called "no seeds"). By comparing the seed-based method to a manually constructed reference set, we can test for coverage issues that stem from collecting reference sets online (versus the posts themselves). Further, this comparison allows us to analyze the trade-off between the high cost in building a manual reference set (versus the low cost of finding seeds) and the gains in accuracy by using the manual reference set. Using "no seeds" tests the effectiveness of constraining the entity tree roots to the seeds. That is, we expect that the "no seeds" method generates a much noisier reference set (leading to poor extraction) than the seed-based method that requires the constraint.

Therefore, for this first experiment, we use the "manual" and "no seed" approaches as baselines to compare the seed-based reference set. For our procedure, for each of our experimental data sets, we build three different reference sets (one manual, one based on seeds, and one without the seeds) and then pass the reference sets to a system that can exploit them to perform automatic extraction (Michelson & Knoblock, 2007). We then compare the extraction results using the standard metrics of recall, precision, and $F_1$-measure. Since the only difference for the extraction algorithm is the reference set provided, the extraction results serve as a proxy for the quality of the reference set both in terms of how well it overlaps with the posts (the coverage) and how clean it is (since noise leads to poor extractions), which is what we want to test.

The goal of our extraction task is to extract the values for given attributes. For instance, using Figure 1, we should extract the model={Accord} and trim={EX}. However, our approach to constructing reference sets does not supply these attribute names. Our method discovers attribute values such as "Honda" and "Accord," but it internally labels their associated attribute names as $attribute_0$ and $attribute_1$, instead of Make and Model. Therefore, to clarify the results we manually label the attribute names as given by the manually constructed reference sets. We do not feel this is much of a hindrance in our method. If a user can find a set of seeds, the user should also be able to find appropriate attributes names. In fact, it is significantly more challenging to discover the attribute values than just finding their names.

### 4.1.1 Data for Seed-based Experiments

We used three real-world data sets as our experimental data. The first set contains classified ads for used cars for sale from the classified site Craigslist.org. Of these, we labeled 1,000 posts to test the extraction of the make (e.g., Honda), model (e.g., Civic), and trim (e.g., DX) attributes. The second set consists of classified ads for used laptops from Craigslist.org as well. Again we labeled 1,000 posts for extracting the manufacturer (e.g., IBM), model (e.g., Thinkpad), and model number (e.g., T41). Our last data set contains posts about skis for sale on eBay. We labeled 1,000 of the posts for extraction of the brand (e.g., Rossignol), the model (e.g., Bandit), and the model specification (e.g., B3, called the "spec"). The data is summarized in Table 6.

We need full, manually constructed reference sets for comparison. For the Cars domain, we collected 27,000 car tuples by pulling data from the Edmunds.com car buying site for cars and combining it with data from a classic car site, SuperLambAuto.com. For the

Table 6: Three experimental data sets

| Name | Source | Attributes | Num. Posts |
|------|--------|-----------|-----------|
| Cars | Craigslist | make, model, trim | 2,568 |
| Laptops | Craigslist | manufacturer, model, model num. | 2,921 |
| Skis | eBay | brand, model, model spec. | 4,981 |

Laptops domain, we scraped 279 laptops off of the online retailer Overstock.com. Lastly, for the Skis domain, we built 213 ski tuples from the skis.com website and cleaned them to remove certain stop words.[5]

The seeds for our seed-based method also came from freely available sources. For the car domain, the seeds consist of 102 car makes, again from Edmunds. The laptop seeds are 40 manufacturers, culled from Wikipedia, and the ski seeds are 18 ski brands pulled from Skis.com.

### 4.1.2 Results for Seed-based Experiments

Table 7 shows the field-level extraction results for each attribute in each domain, comparing the three methods.[6] Again, the "manual" method uses the full reference set constructed from online sources (shown in parentheses in the table), the "no seed" method is the method without seed-based constraint, and our full technique is called "seed-based."

Table 8 summarizes the results, showing the number of attributes where our seed-based method outperformed the other techniques in terms of $F_1$-measure. It also shows the number of attributes where the seed-based technique is within 5% of the other method's $F_1$-measure (including the attributes where the seed-based method outperforms the other method). An $F_1$-measure within 5% is a "competitive" result.

The results show that our seed-based method builds a cleaner reference set than the fully automatic approach that ignores the seeds since the seed-based approach outperforms the "No seed" approach on every single attribute. The seed-based method builds a cleaner, more effective reference set, and that leads to more effective extraction.

The results also support the notion that using the posts themselves to generate a reference set yields reference sets with better coverage than those constructed manually from a single source. Not only does the seed-based method outperform the manual reference sets on a majority of attributes (5/9), the seed-based method's reference set better represents the most specific attributes (ski model, ski model spec., laptop model, and laptop model num.), which are those attributes that are likely to cause coverage problems. For these attributes, only 53.15% of the unique attribute values in the seed-based reference set exist in the manually constructed reference set. Therefore, the coverage is quite different, and given that the seed-based approach performs better on these attributes, its coverage is better.

For example, it is important to note that Overstock sells *new* computers, while the laptops for sale on Craigslist are generally *used, older* laptops. So, while there is a match between the manufacturers (since the laptop manufacturers don't change quickly), even if the used laptops are six months older than the new ones for sale there will be a mismatch between some models and for many of the model numbers. This coverage mismatch using the "manual" reference sets is very clear for the laptop model numbers and ski model specifications. Both of these are attributes that change quite frequently over time as new models come out. This is in contrast to ski brands and laptop manufacturers (our seeds) which change much less frequently and so can be enumerated with less of a concern toward coverage. We note that we chose Wikipedia because of its comprehensive list of laptop manufacturers, but Wikipedia enumerates far fewer models and model numbers than the Overstock reference set and so would be a worse choice for a manual reference set.

---

5. The posts and reference sets for our experiments are available at www.mmichelson.com.

6. Field-level results are strict in that an extraction is correct only if all the tokens that should be labeled are, and no extra tokens are labeled.

Table 7: Extraction results comparing seed-based method

| Cars | | | |
|---|---|---|---|
| *Make* | Recall | Prec. | $F_1$-Meas. |
| Manual (Edmunds) | 92.51 | 99.52 | 95.68 |
| No seed | 79.31 | 84.30 | 81.73 |
| Seed-based | 89.15 | 99.50 | 94.04 |
| *Model* | Recall | Prec. | $F_1$-Meas. |
| Manual (Edmunds) | 79.50 | 91.86 | 85.23 |
| No seed | 64.77 | 84.62 | 73.38 |
| Seed-based | 73.50 | 93.08 | 82.14 |
| *Trim* | Recall | Prec. | $F_1$-Meas. |
| Manual (Edmunds) | 38.01 | 63.69 | 47.61 |
| No seed | 23.45 | 54.10 | 32.71 |
| Seed-based | 31.08 | 50.59 | 38.50 |

| Laptops | | | |
|---|---|---|---|
| *Manufacturer* | Recall | Prec. | $F_1$-Meas. |
| Manual (Overstock) | 84.41 | 95.59 | 89.65 |
| No seed | 51.27 | 46.22 | 48.61 |
| Seed-based | 73.01 | 95.12 | 82.61 |
| *Model* | Recall | Prec. | $F_1$-Meas. |
| Manual (Overstock) | 43.19 | 80.88 | 56.31 |
| No seed | 54.47 | 49.52 | 51.87 |
| Seed-based | 70.42 | 77.34 | 73.72 |
| *Model Num.* | Recall | Prec. | $F_1$-Meas. |
| Manual (Overstock) | 6.05 | 78.79 | 11.23 |
| No seed | 25.58 | 77.46 | 38.46 |
| Seed-based | 34.42 | 86.05 | 49.17 |

| Skis | | | |
|---|---|---|---|
| *Brand* | Recall | Prec. | $F_1$-Meas. |
| Manual (Skis.com) | 83.62 | 87.05 | 85.30 |
| No seed | 60.59 | 55.03 | 57.68 |
| Seed-based | 80.30 | 96.02 | 87.46 |
| *Model* | Recall | Prec. | $F_1$-Meas. |
| Manual (Skis.com) | 28.12 | 67.95 | 39.77 |
| No seed | 51.86 | 51.25 | 51.55 |
| Seed-based | 62.07 | 78.79 | 69.44 |
| *Model Spec.* | Recall | Prec. | $F_1$-Meas. |
| Manual (Skis.com) | 18.28 | 59.44 | 27.96 |
| No seed | 42.37 | 63.55 | 50.84 |
| Seed-based | 50.97 | 64.93 | 57.11 |

Table 8: Summary results of seed-based method versus others

| | Seed vs. No seed | Seed vs. Manual |
|---|---|---|
| Outperforms | 9/9 | 5/9 |
| Within 5% | 9/9 | 7/9 |

Also, we note that our seed-based technique is competitive on 7/9 attributes when compared to the full, manually constructed reference sets. Yet, the number of seeds is drastically smaller than the number of tuples manually constructed for those reference sets. So, even though we are starting with a much tinier set of knowledge, we still retain much of the benefit of that knowledge by leveraging it, rather than having to explicitly enumerate all of the tuple attributes ourselves. This is important as it is much easier to find just the seeds. Therefore, the cost (in manual terms) is much lower for the seed-based approach, but it does not give up accuracy performance as compared to the manual approach for building reference sets.

The one attribute where the manual reference set drastically outperforms our seed-based method is the trim attribute for cars, where the difference is roughly 9% in $F_1$-measure. This is mostly due to the fact that we use field level results, and when the seed-based technique constructs the trim attribute it sometimes leaves out certain attribute tokens. For instance, consider the example where the extracted trim should be "4 Dr DX." Here, the seed-based technique only includes "DX" as the reference set tuple's attribute. Meanwhile, the manually constructed reference set contains all possible tokens since it is scraped from a comprehensive source (its attribute value is "4 Dr DX 4WD"). So, although our seed-based technique finds the DX token and labels it correctly as a trim, it misses the "4 Dr" part of the extraction, so the whole extraction is counted as incorrect using field level results.

Overall, the machine-constructed reference sets yield better extraction results for attributes that occur higher up in the entity trees. The extraction results are best for the attributes at the roots of the entity trees, then the attributes that are children of the roots, and then the leaves of the entity trees. This is largely a discovery issue. The set of possible attribute values generally grows as one traverses the tree (e.g., there are more values for laptop models than manufacturers, and more model numbers than models, etc.). Therefore, the algorithm needs to see more and more posts to overcome the lack of evidence to discover the attributes farther down the entity trees. So, seeing many more posts should generate enough evidence to compensate for this issue.

One limitation of our seed-based technique versus the manual construction of reference sets has to do with the inclusion of certain attributes. Surprisingly, there is not enough repetition in the posts for discovering the years of the cars as attributes. This is due to various factors including the variety of year representations (all four digits, two digits, strange spellings, etc.) and the placement in the posts of the years (since we consider bigrams for subsumptions). However, the full manual reference set does contain years and as such it can extract this attribute, while the seed-based method cannot. Therefore, since our seed-based method was unable to learn how to fit the year attributes into the entity tree, it fails to extract it, and so we remove this attribute from our extraction results (as its results are essentially 0). Nonetheless, although a manual reference set may include an attribute that cannot be discovered automatically, the reference set might have terrible coverage with the posts, limiting its utility. So, we feel it is important to deal with coverage, as our seed-based method does.

### 4.1.3 Results for Iterating for General Tokens

We also tested the effect of iterating to capture "general" tokens versus simply stopping after the first pass over the posts. We again use extraction results as the proxy for comparing these reference sets. In this case, the assumption is that the iterative method will capture more "general" tokens and therefore construct a fuller reference set that yields better extraction results than when the algorithm stops after the first pass over the posts. Table 9 shows the comparable $F_1$-measure results for extraction comparing the "Single Pass" to the "Iterative" approach.

Table 9: Comparing iterating to not iterating (seed-based method)

| Cars | | |
|---|---|---|
| | Single Pass ($F_1$) | Iterative ($F_1$) |
| Make | 93.29 | 94.04 |
| Model | 78.42 | 82.14 |
| Trim | 16.44 | 38.50 |

| Laptops | | |
|---|---|---|
| | Single Pass ($F_1$) | Iterative ($F_1$) |
| Manufacturer | 81.77 | 82.61 |
| Model | 73.52 | 73.72 |
| Model Num. | 49.50 | 49.17 |

| Skis | | |
|---|---|---|
| | Single Pass ($F_1$) | Iterative ($F_1$) |
| Brand | 87.30 | 87.46 |
| Model | 67.03 | 69.44 |
| Model Spec. | 42.75 | 57.11 |

As expected, the iterative technique yields better results. The iterative method outperforms the single-pass approach on every attribute except for one (Laptop Model Numbers, where the $F_1$-measure decreases by -0.33%). Interestingly, when iterating, the algorithm improves more for attributes at deeper levels of the entity trees. That is, when comparing the extraction results for the roots of the entity trees, there is almost no difference. However, at the second level of the entity trees there is a slight improvement (around +3.5% $F_1$-measure for Car Models and +2.5% for Ski Models), and comparing the leaves of the entity trees there is the most improvement (+22% increase for Car Trims, and +15% increase for Ski Model Specifications). So, it seems that the iterating does in fact capture more general tokens which can be used for successful extractions, and it seems that general tokens seem to occur more farther down the trees. We note that the algorithm only iterates a few times for each domain, and since it almost always helps extraction (sometimes quite dramatically), it is a useful component of the seed-based approach to constructing reference sets.

### 4.1.4 Entity Tree Analysis

Although extraction experiments serve as the best metric for the actual utility of the seed-based reference sets, we also ran experiments on the generated entity trees themselves. We examined whether attribute values are consistent in their placement in the entity trees (the column homogeny). For instance, given known car models such as "Civic" we measure if the model values are mostly placed as car model attributes (second level in the tree) or misplaced as car trims (third level). However, measuring this directly without domain expertise is difficult. Instead, we compare the attribute values in the seed-based reference set to those in the manually constructed reference sets, and for those values that match, we measure if they are the for same attribute (i.e., their columns match) or not. This yields a measure of the column homogeny for the seed-based reference set, based on the manual reference set, which is assumed to be clean. However, it is an approximate measure because not all values in the seed-based reference set match those in the manual reference set, since they differ in coverage (see the previous results).

Nonetheless, the results of this approximate measurement indicate a good level of homogeny amongst the seed-based attributes. For skis, only 1.7% of the found attribute values are in the wrong column, while for cars 2.9% of the values are in the wrong columns. Both the skis and cars had a common error of placing the specific attribute (model spec or car trim) one spot higher in the entity tree than it should have been. However, this approximation is misleading for laptops. In the laptops domain, we found perfect column homogeny using this measure, but this is because we can only measure the column homogeny for attributes that match in both the seed-based and manual reference sets. Yet, there were obvious column homogeny errors, such as cpu speeds being placed as model numbers. Since these did not match into the manual reference set, they were ignored by our homogeny measuring experiment. Given that we have enough domain expertise, we did a manual calculation for this set and determined that 8.09% of the tuples in the seed-based set have a cpu speed or other variant as a model number which is incorrect. However, even at 8% this is a good result for homogeny.

### 4.1.5 Comparison Against Supervised Methods

Although our experiments are meant to test the utility of the reference set (not the extraction algorithm itself), one aspect to analyze is reducing the burden on the user for the seed-based approach. That is, by comparing our seed-based reference set (which uses an automatic extractor) to a supervised machine learning approach to extraction, we can examine the amount of labeled data needed for the supervised system to garner similar results. This yields insight into the gain in terms of labor cost because generating the set of seeds is much less costly than labeling data to train a classifier, and we would like to examine how much labeled data is needed to get comparable extraction results.

To analyze the user effort, we compare the seed-based results above to a common machine learning approach for extraction: Conditional Random Fields (CRF) (Lafferty, McCallum, & Pereira, 2001). For this we used MALLET (McCallum, 2002) to implement two different CRF extractors. One, called "CRF-Orth," uses orthographic features of the tokens for extraction, such as capitalization, number containment, etc. The second extractor, "CRF-Win," uses the same orthographic features and also considers a two-word sliding win-dow around a token as a feature. These extractors are built to reflect common features and techniques for CRF-based extraction. Then, we perform 10-fold cross validation for each extractor (varying the amount of the data for training) noting that each fold is independent, and we compare the average field-level extraction results using the supervised approaches to the automatic approach using the seed-based reference set.

Our first experiment compares the seed-based method to each CRF using just 10% of the data for training. This experiment compares the seed-based method to a supervised method using a small enough amount of labeled data to reflect real-world cost constraints. If the seed-based method outperforms the CRFs on a majority of attributes, then it is an effective method for extraction that is also cost effective since it outperforms the supervised methods when they are supplied with a realistic amount of training data. Table 10 shows the summary extraction results for this experiment, similar in format to those of Table 8 where we show the number of times the seed-based method outperforms and is competitive with another method.

Table 10: Summary results comparing the seed-based method to CRFs (10% training data)

|  | Seed vs. CRF-Win | Seed vs. CRF-Orth |
|---|---|---|
| Outperforms | 7/9 | 6/9 |
| Within 5% | 9/9 | 7/9 |

Table 10 shows that our seed-based method outperforms the two CRF extractors on a majority of the attributes, even though the cost in creating a seed list is significantly less than the cost of labeling the data and creating features for training our specific CRFs. Further, the table shows that a technique that relies heavily on structure, such as CRF-Win, performs worse on extraction from posts as compared to other methods.

One aspect to analyze based on these results is the amount of training data needed for the supervised methods to outperform the seed-based methods. For this analysis, we trained the CRFs with 10%, 30%, and 50% of the data, and we note at what amount of training data the CRFs outperform the seed-based method. In certain cases, even with 50% of the data used for training, the CRF did not outperform the seed-based method (we denote this amount as >50% in the table). Table 11 shows the amount of data needed for each CRF to outpeform the seed-based method, broken down by each specific attribute for each domain.

From these results we see that there are quite a few cases when either 50% of the data (or even more) is needed by the supervised approaches to outperform the seed-based method (3/9 for CRF-Orth and 5/9, a majority, for CRF-Win). Therefore, there is a large gain in terms of cost when using the seeds, as so much labeled data would be needed for the supervised systems. In fact, there are attributes where both extractors never outperformed the seed-based approach, even when given 50% of the data for training. Further, we note that only once, for CRF-Orth in the Skis domain, is less than 50% of the data required to outperform the seed-based method for all of the attributes. In this case, using 30% of the data for training is sufficient to outperform the seed-based method on all attribute, but labeling 30% of the data in this domain is far costlier than generating the list of 18 ski brands used as seeds. Lastly, in no domain would just 10% of the data allow the CRFs to outperform the seed-based method on all attributes. Therefore, the seed-based method provides a much less costly approach to this extraction task based on the amount of training data needed.

Table 11: Amount of training data to outperform seed-based approach

| Cars | | |
|---|---|---|
| | CRF-Orth | CRF-Win |
| Make | >50% | >50% |
| Model | 50% | >50% |
| Trim | 10% | 30% |

| Laptops | | |
|---|---|---|
| | CRF-Orth | CRF-Win |
| Manufacturer | 50% | >50% |
| Model | 30% | >50% |
| Model Num. | 10% | 10% |

| Skis | | |
|---|---|---|
| | CRF-Orth | CRF-Win |
| Brand | 30% | 50% |
| Model | 30% | 30% |
| Model Spec. | 10% | 30% |

There is one case where both CRF methods perform well with a relatively small amount of training data, the laptop model number. This attribute fits well for the orthographic features (since it usually has capital letters and numbers in it), and the extractor can generalize well with just 10% training data for extracting it. This argues that certain attributes may benefit from being extracted using generalized features (such as those from CRF-Orth) versus reference-set membership (as our method does). Therefore, we plan to investigate hybrid methods that combine the best of both the CRF-based extractors and reference-set based extractors.

The above set of experiments demonstrate the utility of our seed-based approach. Comparing the seed-based method to a baseline of a manually constructed reference set, we showed that the seed-based method outperforms the manual reference set on a majority of attributes (especially those for which coverage is difficult), while requiring less user effort to construct. Therefore, the reference set constructed using the seeds has better coverage than the manually constructed reference sets and can be used effectively, even though it is cheaper to construct. Further, we showed that the constraint that our seed-based method uses (constraining the roots to be seeds) does indeed have a strong impact on the results, versus not using this constraint. Lastly, we compared using a seed-based approach to a supervised machine learning approach to judge the comparable amount of training data needed to outperform the seed-based method, and we show that indeed, it takes quite a bit of training data as compared to the small number of seeds.

## 4.2 Experiments: Locking Approach

Our next set of experiments analyze our locking-based approach to building reference sets. As stated above, the locking based technique is appropriate for the special case where seeds are too costly or impossible to find. Therefore, our locking based approach is an alternative to the "no seed" method described in our previous experiments. The experimental procedure for these experiments is the exactly same as above. We use the same data sets and compare the reference sets constructed in different ways (seed-based, "no seed," and "locked") by passing them to the same extraction mechanism and comparing the extraction results as a proxy for the reference set's utility.

For the locking algorithm, we must specify the number of posts to add at each locking iteration. We set this value to 200, which is large enough to limit the total possible number of iterations (versus say, 20), but also small enough to allow the algorithm to converge before seeing all of the posts (versus, say 1,000, which may be too coarse). Table 12 shows the total number of posts (and iterations) required for the locking algorithm to lock all levels of the entity trees, and therefore converge, returning the constructed reference set. We note that for all domains, the total number of posts required for the locking algorithm to converge was less than the total number of posts for that domain. Table 12 also shows the total number of posts for each domain.

Table 12: Locking convergence results

| Domain | Total Posts Required for Locking | Total Possible Posts | Iterations |
|--------|-----------------------------------|----------------------|------------|
| Cars | 2,000 | 2,568 | 10 |
| Laptops | 2,400 | 2,921 | 12 |
| Skis | 4,400 | 4,981 | 22 |

Table 13 shows the comparative field-level extraction results for the different domains, using the reference sets generated by the different methods (seed-based, "no seed," and "locked").

Based upon the results of Table 13, we see that locking is a good alternative to simply not locking at all ("no seeds"). For the cars and the skis domains, there is only one difference in $F_1$-measure that is statistically significant using a two-tailed test at 95% confidence (the Car Trim attribute). However, in the laptops domain, the locking method outperforms "no seed" on all of the attributes, largely due to its increase in precision, which is a direct result of locking noise out of the entity trees. Therefore, it is a good strategy to attempt the locking approach (versus the "no seed" approach) since based upon these results, in the worst case, it will have a minimal negative effect on the generated reference set (evidenced by the Car Trim attribute), but it can yield significantly cleaner reference set in the best case (as shown by the laptop results). Further, the locking algorithm converged (i.e., was able to lock all levels before it saw all posts) in all domains, so it was able to produce reference sets without the burden of requiring additional posts.

Table 13: Extraction results comparing locking method

| Cars | | | |
|---|---|---|---|
| *Make* | Recall | Prec. | $F_1$-Meas. |
| Locked | 79.64 | 84.46 | 81.84 |
| No seed | 79.31 | 84.30 | 81.73 |
| Seed-based | 89.15 | 99.50 | 94.04 |
| *Model* | Recall | Prec. | $F_1$-Meas. |
| Locked | 65.30 | 83.24 | 72.22 |
| No seed | 64.77 | 84.62 | 73.38 |
| Seed-based | 73.50 | 93.08 | 82.14 |
| *Trim* | Recall | Prec. | $F_1$-Meas. |
| Locked | 19.54 | 52.13 | 28.28 |
| No seed | 23.45 | 54.10 | 32.71 |
| Seed-based | 31.08 | 50.59 | 38.50 |

| Laptops | | | |
|---|---|---|---|
| *Manufacturer* | Recall | Prec. | $F_1$-Meas. |
| Locked | 60.42 | 74.35 | 66.67 |
| No seed | 51.27 | 46.22 | 48.61 |
| Seed-based | 73.01 | 95.12 | 82.61 |
| *Model* | Recall | Prec. | $F_1$-Meas. |
| Locked | 61.91 | 76.18 | 68.31 |
| No seed | 54.47 | 49.52 | 51.87 |
| Seed-based | 70.42 | 77.34 | 73.72 |
| *Model Num.* | Recall | Prec. | $F_1$-Meas. |
| Locked | 27.91 | 81.08 | 41.52 |
| No seed | 25.58 | 77.46 | 38.46 |
| Seed-based | 34.42 | 86.05 | 49.17 |

| Skis | | | |
|---|---|---|---|
| *Brand* | Recall | Prec. | $F_1$-Meas. |
| Locked | 60.84 | 55.26 | 57.91 |
| No seed | 60.59 | 55.03 | 57.68 |
| Seed-based | 80.30 | 96.02 | 87.46 |
| *Model* | Recall | Prec. | $F_1$-Meas. |
| Locked | 51.33 | 48.93 | 50.10 |
| No seed | 51.86 | 51.25 | 51.55 |
| Seed-based | 62.07 | 78.79 | 69.44 |
| *Model Spec.* | Recall | Prec. | $F_1$-Meas. |
| Locked | 39.14 | 56.35 | 46.29 |
| No seed | 42.37 | 63.55 | 50.84 |
| Seed-based | 50.97 | 64.93 | 57.11 |

## 4.3 Experiments: Assumptions for Constructing Reference Sets

In this section we examine some of the assumptions made on the data from which we are able to construct the reference sets using the seed-based method (or the locking method, if necessary). First, we mentioned previously that we do not need to assume that the constructed reference set is filled only with single token attributes, as roughly 6% of the attributes in the constructed reference sets are n-grams using the seed-based method. We note, however, that other researchers have also discussed the difficulty in discovering n-grams for concept hierarchies from text. For example, some previous work discards concepts from the topic hierarchy if they consist of multiple terms (Bast et al., 2006).

Despite the fact that our merging heuristic can yield n-gram values for attributes, there are cases where the heuristic breaks down. Specifically, it does not perform well in domains where most of the attribute values are multiple tokens, and those tokens occur with each other in various frequencies. For example, consider if users are selling items for sports teams, such as a "San Diego Chargers helmet," and a "San Francisco 49rs t-shirt." In this case, the "Diego" and "Francisco" tokens will be subsumed by "San" creating an unnecessary entity tree, rather than joining both of the terms together. This is the main failure of our merging approach: the algorithm will force a subsumption relationship because the merging rule fails to fire. However, we note that our merging heuristic never causes the subsumption rule not to fire, rather it fails to do so itself and therefore creates errant entity trees. Therefore, since it never destroys information for an entity tree (e.g., never causes subsumption to fail) we consider two approaches to improving the merge heuristic. First, we could make the approach more aggressive in merging. Second, we could perform some post-hoc analysis of the constructed entity trees to fix these errors. For instance, one could use outside information, such as an ontology or corpus statistics, to determine the likelihood of Franscisco being a child of San versus being merged into a single term and therefore cleaning up the hierarchy to account for failed merging. Improving our merging method is a future research challenge.

While it might seem that it is necessary for the attributes to always appear adjacent to one another in the posts (e.g., "make model trim" with no other tokens between them), this is not the case. In fact, on average, across the three domains we measured 0.115 tokens in between each attribute in the posts. This is significantly larger than 0, and it implies that indeed there are tokens between the attributes in a number of the posts. Further, it might seem that we must always see the correct order in the posts (e.g., never see a Car trim attribute before a Car make attribute) in order to successfully construct the reference sets. However, again, due to the unstructured nature of the posts, we see that this is not the case. In fact, we only see "perfect ordering" for less than half of the posts in the three domains (45.2%), where perfect ordering is the full ordering defined by the entity tree (e.g., a Car post with a make attribute a model attribute and a trim attribute, in that order). In fact, in cars 0.61% of the posts have the Car trim attribute coming before any of the other attributes, which is a strange and random ordering. Therefore, we do not need to assume that the ordering is always correct or that there will not be extraneous tokens in between the attributes. And we do not need to assume that each attribute is a single token. This is because there are enough cases where the ordering does reflect the entity tree, and words are seen together, such that the subsumption heuristic can fire to produce a correct entity tree.

Yet, there are a certain number of assumptions we do make in order to construct a reference set. First, we must assume that there is some reasonable hierarchical structure for the entity trees (e.g., Car makes are more general than Car models). Although this constraint holds for an enormous set of categories (e.g., all items for sale, descriptive categories such as geographical and person data, etc.) there are some categories that lack this characteristic (e.g., personal ads). We see that this is the case when we analyze information about hotels, using the "Bidding For Travel" extraction data set from previous work (Michelson & Knoblock, 2008). In this data set, the goal is to extract hotel information, such as hotel names and local areas from posts to an internet forum where users talk about their deals they received for hotel accommodations. When we give our seed-based method the set of hotel names as seeds (from the data set) and try to build a reference set from the roughly 2,500 posts from BiddingForTravel.com, we expect it to construct entity trees with the local areas as children of the hotel names. However, the resulting reference set is not constructed well. In fact, the algorithm only finds 20 hotel tuples out of the 132 possible. This is mostly due to the fact that even the users themselves, who created these posts, cannot decide on a consistent hierarchical structure for this data. Analyzing these posts, users put the hotel name immediately before the local area 40.58% of the time and the local area immediately before the hotel name 27.17% of the time. Therefore, even to the users themselves there are least two intepretations for representing the data as entity trees, and it is ambiguous as to whether we should expect the hotel names as roots or the areas as roots. We note that it is often the case that for the same hotel the users flip the mentions of these attributes. So, the assumption that we do make is not that the posts themselves are structured, but that the structure of the entity trees is consistent (and agreed upon) such that the algorithm can reconstruct the entity trees based on the users' posts. That is, for this case, all entity trees should be rooted on either hotel names, or local areas, but not a mixture as we see reflected by the users' posts. Further, this data is particularly difficult for constructing reference sets because the terms for the attributes are freely intermixed. That is, tokens from a hotel name and the local area are sometimes interspersed, which makes it difficult for the machine to determine which terms go together in an attribute. This is largely due to our limiting of bi-grams to being "in order" (which we do for efficiency), and perhaps if we extend the algorithm to consider all possible combinations of bigrams (perhaps by extending our method to a distributed approach) we could handle this issue.

## 5. Related Work

The focus of this research is in creating reference sets from posts. As the reference sets are flattened entity trees, the work that most closely resembles ours is the research on creating term hierarchies (subsumptions) from text. There are alternative methods for building term hierarchies. However, these methods are not as well suited as the Sanderson and Croft method (1999) that we chose for our problem. First, since our data is ungrammatical, we cannot use subsumption methods that rely on Formal Concept Analysis, which relate verbs to nouns in the text to discover subsumptions (Cimiano, Hotho, & Staab, 2005). We have plenty of nouns in posts, but almost no verbs. Further, since our algorithm runs iteratively due to the "general token" problem, we need a method that runs efficiently. Our algorithm using the Sanderson and Croft method runs in $O(kn)$ time where $n$ is the number of tokens from the posts, and $k$ is the number of iterations for general tokens, since our process scans the posts to create the bigrams and calculate the probabilities and then considers only those with high enough probabilities. This is in contrast to other methods for term subsumption that use Principle Component Analysis (PCA) (Dupret & Piwowarski, 2006; Bast et al., 2006) and run in $O(n^3)$ time with respect to the token-by-token matrix (which may be sparse). Therefore, these PCA methods are not suitable to a large number of tokens and for more than one iteration. There is also previous work that uses outside information sources, such as links between image tags and the users who supply them on Flickr, to aid in the building of term hierarchies (Schmitz, 2006). We do not have explicit links between terms and users. Lastly, there is previous work that uses Google to determine term dependencies (Makrehchi & Kamel, 2007). However, we cannot assume that the terms we encounter in posts will occur across many webpages. In fact, our method will work even if the posts were the only place that mentions the entities on the entire Web (provided there are enough posts). However, for such a case, which might occur if the posts are for obscure items, we could not leverage their Google-based method. Further, although there are a few alternative approaches to building term hierarchies, there is also one larger, more fundamental difference between our problem and the previous work on ontology creation.

All of the previous methods build up a single, monolithic *conceptual* hierarchy. In our case, we instead aim to build a number of disjoint entity trees, which we can then flatten into a reference set.

Along the lines of discovering term hierarchies from text, there is also work that aims to extract features from text such as product reviews. Since the entity trees our method constructs often include product features (e.g., Laptop models), our work constructs similar output to these methods. Approaches to feature extraction include association rule mining for finding frequent product features (Hu & Liu, 2004) and leveraging the Web to aid in feature extraction (Popescu & Etzioni, 2005). However, these methods rely on natural language processing, such as part-of-speech (POS) tagging to parse the reviews into possible features. However, our posts are not grammatical enough to support part-of-speech tagging, and so we cannot use such features of the data.

Recently, even some of the large commercial search engines have begun to construct reference sets from Web data. Google has built "Google Squared,"[7] which allows users to type a query category and returns a "square" which is a reference set related to the category. This product provides an intuitive interface for including/excluding attributes (columns) and naming them appropriately. However, when we tested the application for our experimental domains (providing queries of "cars," "laptops," and "skis") we found that the columns in the square were not broken down into the constituent attributes as finely as our reference sets. For instance, for both laptops and skis, there is a single attribute (called "item name") that essentially functions as a post describing the entity. For laptops, the "item name" combines the manufacturer, model, and model number into a single attribute, and for skis it combines the brand, model, and model spec. For cars, the square sometimes combined the make and model of the car into the "item name." Yet, it is encouraging that such a large company finds reference sets important and useful enough to devote an application to them, and therefore we feel that methods such as ours could greatly complement this technology by providing a means to build even finer grained reference sets to include in the "squares."

As stated, the goal of this work is to build reference sets, which can be used in a number of tasks, including ontology maintenance, query formulation, and information extraction. Given that we chose information extraction as the mechanism for evaluating our reference sets, for completeness we describe related work on information extraction to put the application of our constructed reference sets in context. We also point readers to the previous work on reference-set-based information extraction, which also present comparisons to other extraction methods (Michelson & Knoblock, 2005, 2007, 2008, 2009).

We note that there are information extraction techniques that are based on CRFs that directly use reference sets in the form of either single columns of a reference set ("dictionaries") (Cohen & Sarawagi, 2004) or full, relational databases (Mansuri & Sarawagi, 2006), and in fact, these approaches focus on extraction from unstructured text, similar to posts. Again, our work in this paper complements these methods well as either a constructed reference set can be split column-wise to produce dictionaries, or the reference-set itself can be used as the relational database.

---

7. www.google.com/squared

Similarly to the reference-set-based extraction methods described above, there are extraction methods that use ontologies as their background information (Embley, Campbell, Jiang, Liddle, Lonsdale, Ng, & Smith, 1999). Later versions of their work even talk about using ontology-based information extraction as a means to semantically annotate unstructured data such as car classifieds (Ding, Embley, & Liddle, 2006). Although the ways in which the methods use the background information differ (the ontology-based method performs keyword-lookup into the ontology along with structural and contextual rules, while the reference-set-based methods use either machine-learning or string-similarity methods to match posts to reference-set members), our reference-set construction method presented here can complement this ontology-based extraction well. One difficulty with an ontology-based method is that creating and maintaining an ontology is an expensive data engineering task. Perhaps reference-set construction methods such as this one can ease this burden by providing a method that can discover ontology instances and automatically map their relations. Along these lines are methods that use informal ontologies, such as Wikipedia, as their background information (Wu, Hoffmann, & Weld, 2008; Kazama & Torisawa, 2007). Again, our method is complementary here, especially as it is unclear whether Wikipedia would cover some of the more obscure tuples we could generate for our reference set. In fact, as we showed in the Laptop domain, the Wikipedia source of laptops was not nearly as comprehensive as those from Overstock, which themselves did not appropriately cover the posts in the same way our generated reference set did.

While there are previous approaches that use outside information to aid extraction (such as ontologies and reference sets), there are also quite a few unsupervised methods for extraction. Although they do not use reference sets, we include them here for completeness in addressing the extraction problem. One set of these techniques focuses on finding relations from the Web. Aggregating these relations can sometimes yield reference sets, such as a table of person $X$ "being born in country" $Y$ (Cafarella, Downey, Soderland, & Etzioni, 2005; Hassan, Hassan, & Emam, 2006; Pasca, Lin, Bigham, Lifchits, & Jain, 2006). However, this research differs from reference-set-based methods for posts because they extract such relations from Web pages which allows them to learn and exploit specific extraction patterns. These patterns assume that similar structures will occur again to make the learned extraction patterns useful, but such structural assumptions about posts cannot be made beyond the redundancy of bigrams. Again, however, such work complements ours quite well, as both methods could perhaps be used together to build reference sets from both Web pages and posts.

We note that the task of extracting information from posts that may not have an associated reference set (e.g., apartment listings, for which an entity tree would be hard to define) has received attention in the past as well. So, the space of extraction from posts both with and without reference sets is covered by complementing the previous work on reference-set-based extraction from posts with this other work on extraction from posts. In particular, some previous work uses information about the structure of the ads to perform extraction (e.g., the idea that for certain types of posts attributes are often multi-token) (Grenager, Klein, & Manning, 2005). More similar to our work is that which uses "prototype learning" for extraction from posts where seed examples for each of the attributes to extract, called "prototypes," are provided as background knowledge (Haghighi & Klein, 2006). Our machine-constructed reference sets could provide these prototypes automating such an approach even further. One interesting approach that can be used for both posts that have an associated reference set and those that do not was presented by Chang, et. al. (2007), where the extraction algorithm encodes and uses various constraints for extraction. One such supported constraint is dictionary membership. As we described previously, such dictionaries can be built directly from the reference sets that our approach constructs.

## 6. Conclusion

This paper presents a method for constructing reference sets from unstructured, ungrammatical text on the Web. Once discovered, these reference sets can be used for tasks including ontology maintenance, query formulation, and information extraction. We demonstrate the utility of the machine constructed reference-sets by comparing them to manually constructed reference sets in an information extraction task, and we show that the machine constructed reference sets yield better extraction results.

In the future we plan to investigate synonym discovery and its relationship to automatically constructing reference sets. For instance, we may be able to automatically merge branches of the hierarchy if they are synonyms referring to the same object (such as "Lenovo" and "IBM" laptops). Further, we plan to investigate the topic of dynamic data integration using automatically mined reference sets. That is, once the system discovers a reference set, we would want it to bring in other related sources for data integration.

## Acknowledgments

This research is based upon work supported in part by the National Science Foundation under award number CMMI-0753124, in part by the Air Force Office of Scientific Research under grant number FA9550-07-1-0416, and in part by the Defense Advanced Research Projects Agency (DARPA), through the Department of the Interior, NBC, Acquisition Services Division, under Contract No. NBCHD030010.

The U.S. Government is authorized to reproduce and distribute reports for Governmental purposes notwithstanding any copyright annotation thereon. The views and conclusions contained herein are those of the authors and should not be interpreted as necessarily representing the official policies or endorsements, either expressed or implied, of any of the above organizations or any person connected with them.